\pdfoutput=1
\documentclass[11pt]{article}

\usepackage[]{ACL2023}

\usepackage{times}
\usepackage{latexsym}

\usepackage{pdflscape}
\usepackage{soul}
\usepackage{caption}
\usepackage{amsmath}
\usepackage{amsthm}
\usepackage{booktabs}
\usepackage{algorithm}
\usepackage{algorithmic}
\usepackage{array,multirow,graphicx}
\usepackage{xcolor} 
\usepackage{makecell}
\usepackage{soul}
\usepackage{tikz}
\usetikzlibrary{positioning,fit,arrows.meta,backgrounds}
\usepackage{graphicx}
\usepackage{booktabs}
\usepackage{tikz}
\tikzset{
    module4/.style={%
        draw, rounded corners,
        minimum width=55mm,
        align=center,
        minimum height=15mm,
        font=\sffamily
        },
    module/.default=2cm,
    >=LaTeX
}
\tikzstyle{pinstyle} = [pin edge={<-,ultra thick,black}]
\tikzstyle{pinstyle-1} = [pin edge={->,ultra thick,black}]
\tikzstyle{pinstyle-2} = [pin distance=9cm, edge={->,ultra thick,black}]

\usepackage[T1]{fontenc}
\usepackage[utf8]{inputenc}

\usepackage{microtype}

\usepackage{inconsolata}

\title{Deep Learning Approaches to Lexical Simplification: A Survey}

\author{
    Kai North\textsuperscript{1}, Tharindu Ranasinghe\textsuperscript{2}, Matthew Shardlow\textsuperscript{3}, Marcos Zampieri\textsuperscript{1} \\
    \textsuperscript{1}George Mason University, USA,
   \textsuperscript{2}Aston University, UK \\ \textsuperscript{3}Manchester Metropolitan University, UK \\
    knorth8@gmu.edu
}

\begin{document}

\maketitle

\begin{abstract}
 Lexical Simplification (LS) is the task of replacing complex for simpler words in a sentence whilst preserving the sentence's original meaning. LS is the lexical component of Text Simplification (TS) with the aim of making texts more accessible to various target populations. A past survey \cite{PaetzoldSpecia2017_surveyLS} has provided a detailed overview of LS. Since this survey, however, the AI/NLP community has been taken by storm by recent advances in deep learning, particularly with the introduction of large language models (LLM) and prompt learning. The high performance of these models sparked renewed interest in LS. To reflect these recent advances, we present a comprehensive survey of papers published between 2017 and 2023 on LS and its sub-tasks 
 %(substitute generation, substitute selection, and substitute ranking) 
 with a special focus on deep learning. We also present benchmark datasets for the future development of LS systems.
\end{abstract}

\section{Introduction}

%Lexical Simplification (LS) is a Natural Language Processing (NLP) task. 
LS improves the readability of any given text with the aim of helping vocabulary and literacy development. LS achieves this by replacing complex words in a sentence with simpler alternatives. LS returns a simplified sentence which can be passed to a TS system for further syntactic and grammatical simplification. The replaced complex words are those words which a general or targeted population found to be hard to read, interpret, or understand. Previous LS systems have been designed to simplify complex words for children, second language learners, individuals with reading disabilities or low-literacy \cite{PaetzoldSpecia2017_surveyLS}. LS therefore provides both developers and users with a degree of personalization that is unattainable through seq2seq or generative TS systems \cite{yeung-lee-2018-personalized,Lee_Yeung2018}. 

Deep learning, and latterly, LLM and prompt learning, have revolutionized the way we approach many NLP tasks, including LS. Previous LS systems have relied upon lexicons, rule-based,  statistical, n-gram, and word embedding models to identify and then simplify complex words \cite{PaetzoldSpecia2017_surveyLS}. These approaches would identify a complex word, for example,  ``\textit{bombardment}'' as being in need of simplification and would suggest ``\textit{attack}''  as a suitable alternative (Figure \ref{pipeline_figure}), hereby referred to as a candidate substitution. 

State-of-the-art deep learning models, such as BERT \cite{devlin2019bert}, RoBERTa \cite{liu2019roberta}, GPT-3 \cite{gpt_3}, and others, automatically generate, select, and rank candidate substitutions with performances superior to traditional approaches. These include relying on pre-existing lexicons, simplification rules, or engineered features \cite{tsar2022}. There have been no surveys published on deep learning approaches for LS. The paper by \citet{PaetzoldSpecia2017_surveyLS} is the most recent survey on LS but it precedes studies that demonstrate the headway made by state-of-the-art deep learning approaches. A broad comprehensive survey on TS was published in 2021\cite{TSsurvey_thanyyan_et_al}. However, this survey likewise does not cover recent advances in the field nor does it focus specifically on LS. This paper therefore continues pre-existing literature by providing an updated survey of the latest deep learning approaches for LS and its sub-tasks of substitute generation (\textbf{SG}), selection (\textbf{SS}), and ranking (\textbf{SR}). 

% The survey has the following format. Sections \ref{pipeline} and \ref{eval} detail the LS pipeline and its evaluation metrics respectively. Section \ref{datasets} provides a list of LS datasets and resources that can be used by deep learning approaches. Throughout Section \ref{deep_learning}, we discuss the progress made in all tasks of LS since 2017. Section \ref{conclusion} summarizes with a brief discussion on the future of LS research. 

\section{Pipeline}\label{pipeline} % Could be a subsection of introduction. 

We structure this survey around the main components of the LS pipeline: SG, SS, and SR (Section \ref{deep_learning}). We also provide an overview of recent datasets (Section \ref {datasets}), and discuss open challenges in LS  (Section \ref {challenges}). Normally, an LS pipeline starts with complex word identification (\textbf{CWI}). However, since it is often considered as a standalone precursor, we refer the reader to \citet{LCPsurvey}, for a detailed survey on CWI methods.

% \paragraph{\bf{Complex Word Identification}} There are two approaches to CWI. The first approach is a binary classification task. Target words are assigned with a non-complex (0) or complex (1) label \cite{paetzold-specia:2016:SemEval1,yiman-EtAl:2018:BEA}. The second approach is a regression-based task, referred to as Lexical Complexity Prediction (LCP). LCP assigns target words with a value between 0 and 1. Gold labels are obtained via lickert-scale annotation \cite{maddela2018word,shardlow-etal-2020-complex}. Typically, a 5 or 6-point licket-scale is used providing labels ranging from very simple (1), neutral (3), to very complex (5) \cite{shardlow-etal-2020-complex}. LCP is considered superior to binary CWI \cite{LCPsurvey}. It allows for the prediction of a neutral level of complexity and as a result, is able to classify words on the decision boundary (words which have been labeled as being both non-complex and complex).

\begin{figure}[!ht]
  \centering
\scalebox{0.67}{\begin{tikzpicture}
\node[module4, fill=white!5] (I1) {\textbf{Complex Sentence}\\\\ Bombardment by regime forces};
\node[module4, below=0.75cm of I1] (I2) {\textbf{Complex Word Identification}\\\\ \textbf{CWI}: \underline{\textit{Bombardment}}};
\node[module4, fill=white!5, below=0.75cm of I2] (I3) {\textbf{Substitute Generation}\\\\ \textbf{SG}: assault, raid, attack};
\draw[->, ultra thick] (I1)--(I2);
\draw[->, ultra thick] (I2)--(I3);

\node[module4, fill=white!5, right=0.5cm of I1] (I6) {\textbf{Simplified Sentence}\\\\ Attack by regime forces};
\node[module4, fill=white!5, below=0.75cm of I6] (I5) {\textbf{Substitute Ranking}\\\\  \textbf{SR:} \#1. \underline{\textit{attack}}, \#2. assault};
\node[module4, fill=white!5, below=0.75cm of I5] (I4) {\textbf{Substitute Selection}\\\\ \textbf{SS}: assault, \st{raid}, attack};
\draw[->, ultra thick] (I3)--(I4);
\draw[->, ultra thick] (I4)--(I5);
\draw[->, ultra thick] (I5)--(I6);
\end{tikzpicture}}
\caption{LS Pipeline. SG, SS, and SR are the main components of LS.}\label{pipeline_figure}
\end{figure}

% LS Pipeline. SG, SS, and SR are the main components of LS and are the sub-tasks of interest in this survey. CWI is often considered a separate precursor

%\vspace{-3mm}

\paragraph{\bf{Substitute Generation}} SG returns a number: \textit{k}, of candidates substitutions that are suitable replacements for a previously identified complex word. Usually, an LS system will generate candidate substitution in the range of \textit{k} = [1, 3, 5, or 10] with \textit{top-k} referring to the most appropriate candidates. These candidate substitutions need to be more simple, hence easier to read, interpret, or understand than the original complex word. The candidate substitutions also need to preserve the original complex word's meaning, especially in its provided context.

\paragraph{\bf{Substitute Selection}} SS filters the generated \textit{top-k} candidate substitutions and removes those which are not suitable. For instance, candidate substitutions which are not synonymous to the original complex word or that are more complex are often removed.

\paragraph{\bf{Substitute Ranking}} SR orders the remaining \textit{top-k} candidate substitutions from the most to the least appropriate simplification. The original complex word is then replaced with the most suitable candidate substitution.

\subsection{Evaluation Metrics}\label{eval} 

 % A true positive is a generated top-\textit{k} candidate substitution that belongs within a corresponding number of top gold labels, whereas a false negative is a generated top-\textit{k} candidate substitution that does not. A false positive is a gold label that is not within a returned number of top-k candidate substitutions. A true negative is a candidate substitution that was correctly removed during SS or not returned during SG.

 All sub-tasks of the LS pipeline are evaluated using precision, accuracy, recall, and F1-score. Several additional metrics have also been used: potential, mean average precision (MAP), and accuracy at top-\textit{k}.  Potential  is the ratio of predicted candidate substitutions for which at least one of the top-\textit{k} candidate substitutions generated was among the gold labels \cite{tsar2022}. MAP evaluates whether the returned top-\textit{k} candidate substitutions match the gold labels as well as whether they have the same positional rank. Accuracy at top-\textit{k} = [1, 2, or 3] is the ratio of instances where at least one of the candidate substitutions at \textit{k} is among the gold labels.
 
\section{Deep Learning Approaches}\label{deep_learning}

\begin{table*}[!ht]
\centering
\scalebox{0.7}{\begin{tabular}{cc|c|c|ccc|c}

% \multicolumn{1}{c}{\textbf{}} & \multicolumn{1}{c}{\textbf{}} \multicolumn{8}{c}{}\\
\hline
       \multicolumn{2}{c|}{\textbf{Deep Learning Approaches}} & \textbf{ACC} & \textbf{ACC@1}  & \textbf{ACC@3} & \textbf{MAP@3} & \textbf{Potential@3} & \textbf{Paper}  \\
      \hline
         \multicolumn{1}{c}{SG} & \multicolumn{1}{c}{SS \& SR} & \multicolumn{5}{c}{TSAR-2022 (EN)} & \multicolumn{1}{c}{} \\
      \hline
       GPT-3+Prompts & GPT-3 &  \textbf{0.8096} & \textbf{0.4289} & \textbf{0.6863 }& \textbf{0.5834} &\textbf{0.9624} & \cite{unihd-tsar-2022-shared-task}  \\
       MLM & LLM+Embeddings+Freq &   0.6568 & 0.3190 & 0.5388  & 0.4730 & 0.8766 & \cite{mantis-tsar-2022-shared-task}  \\
       LLM+Prompt & MLM Prediction Score&  0.6353 & 0.2895 & 0.5308 & 0.4244 & 0.8739 & \cite{uom-mmu-tsar-2022-shared-task}  \\
     %   MLM & MLM Prediction Score&   0.5978 & 0.3029 & 0.4079  & 0.8230 & \cite{qiang2020BERTLS}  \\
     %  LM+Embeddings & POS+LM+Embeddings &   0.5442 & 0.2359 & 0.3823 & 0.8310 & \cite{rcml-tsar-2022-shared-task}  \\
      %  MLM & MLM Prediction Score&   0.5174 & 0.2493 & 0.3522 & 0.7533 & \cite{gmu-wlv-tsar-2022-shared-task} \\
      % MLM+Others$^1$ & MLM Prediction Score&   0.4664 & 0.1823  & 0.2743 &  0.6729 & \cite{teampn-tsar-2022-shared-task} \\
      %  MLM & LM+GPT-2+Embeddings &   0.4316 & 0.2064 &  0.1995 & 0.6139 & \cite{polyu-cbs-tsar-2022-shared-task} \\
      % MLM & Embeddings+POS & 0.4021 & 0.1581 & 0.2603 & 0.6568 &  \cite{presiuniv-tsar-2022-shared-task} \\
      % LM+Embeddings & LM+Embeddings & 0.3860 & 0.1957 & 0.2603 &  0.5656 &  \cite{cils-tsar-2022-shared-task} \\
      %  MLM & Freq+BinaryClassifier & 0.3619 & 0.1152 & 0.2573 &  0.6541 & \cite{cental-tsar-2022-shared-task}\\
      \hline
         \multicolumn{1}{c}{SG} & \multicolumn{1}{c}{SS \& SR} & \multicolumn{5}{c}{TSAR-2022 (ES)} & \multicolumn{1}{c}{} \\
      \hline
       GPT-3+Prompts & GPT-3 &  \textbf{0.6521} & \textbf{0.3505} & \textbf{0.5788} & \textbf{0.4281 } &\textbf{0.8206} & \cite{unihd-tsar-2022-shared-task} \\
       MLM & Embeddings+POS &  0.3695 & 0.2038 & 0.3288 &  0.2145 & 0.5842 &  \cite{presiuniv-tsar-2022-shared-task} \\
       LLM+Prompt & MLM Prediction Score&  0.3668 & 0.160 & 0.2690 & 0.2128 & 0.5326 & \cite{uom-mmu-tsar-2022-shared-task}  \\      
      %  MLM & LM+GPT-2+Embeddings &  0.3586 & 0.1630 &  0.2068 & 0.5244 & \cite{polyu-cbs-tsar-2022-shared-task} \\
      %  MLM & MLM Prediction Score& 0.3532 & 0.1820 & 0.2202 & 0.5679 & \cite{gmu-wlv-tsar-2022-shared-task} \\
       % MLM & Freq+BinaryClassifier &  0.3097 & 0.1467 & 0.1826 &  0.5000 & \cite{cental-tsar-2022-shared-task}\\   
       % MLM & MLM Prediction Score&  0.2880 & 0.0951 & 0.1868 & 0.4945 & \cite{qiang2020BERTLS}  \\
        \hline
         \multicolumn{1}{c}{SG} & \multicolumn{1}{c}{SS \& SR} & \multicolumn{5}{c}{TSAR-2022 (PT)} & \multicolumn{1}{c}{} \\
      \hline
         GPT-3+Prompts & GPT-3 &  \textbf{0.7700} & \textbf{0.4358} & \textbf{0.6299} & \textbf{0.5014} &\textbf{0.9171} & \cite{unihd-tsar-2022-shared-task} \\
       MLM & MLM Prediction Score& 0.4812 & 0.2540 & 0.3957 & 0.2816 & 0.6871 & \cite{gmu-wlv-tsar-2022-shared-task} \\
       MLM & Freq+BinaryClassifier & 0.3689 & 0.1737 & 0.2673 & 0.1983 &  0.5240 & \cite{cental-tsar-2022-shared-task}\\   
      %  MLM & LM+GPT-2+Embeddings &  0.3262 & 0.1390 &  0.1755 & 0.4491 & \cite{polyu-cbs-tsar-2022-shared-task} \\
       % MLM & MLM Prediction Score& 0.3262 & 0.1577 & 0.1904 & 0.4946 & \cite{qiang2020BERTLS}  \\       
       % MLM & Embeddings+POS & 0.3074 & 0.1604 &  0.1573 & 0.4598 &  \cite{presiuniv-tsar-2022-shared-task} \\
       % LM+Prompt & LM prediction & 0.1711 & 0.0695  & 0.1011 & 0.2486 & \cite{uom-mmu-tsar-2022-shared-task}  \\  
     \hline

\end{tabular}}
 \caption{The top 3 deep learning approaches across the TSAR-2022 datasets. Best performances in bold.}\label{results_deeplearning}
%  {\label{tab:table-name}target words are in bold.}
\end{table*}

% $^1$\textit{Others} refers to VerbNet, PPDB, and Knowledge Graph (Section \ref{SR}).

Prior to deep learning approaches, lexicon, rule-based, statistical, n-gram, and word embedding models were state-of-the-art for SG, SS, and SR. As previously mentioned, \citet{PaetzoldSpecia2017_surveyLS} have provided a comprehensive survey detailing these approaches, their performances, as well as their impact on LS literature. The following sections provide an extension of the work carried out by \citet{PaetzoldSpecia2017_surveyLS}. We introduce new deep learning approaches for LS and begin our survey of the LS pipeline at the SG phase. The recent developments in the CWI step of the pipeline have been extensively surveyed by \citet{LCPsurvey}.

% \subsection{Complex Word Identification}\label{CWI}

\subsection{Substitute Generation}\label{SG}

% \cite{Glavas2015}'s model is referred to as LIGHT-LS 
% \paragraph{\bf{Word Embeddings}} Word embeddings have been used for SG by \cite{Glavas2015} and \cite{paetzold-specia-2017-lexical}. \cite{Glavas2015} employed GloVe \cite{} word embeddings, whereas  \cite{paetzold-specia-2017-lexical} used a context-aware word embeddings model trained over a corpus annotated with POS tags and retrofitted with synonym relations per \cite{}. Both word embeddings models conducted SG by selecting those word embeddings with the highest cosine similarity \cite{Glavas2015}, or lowest cosine distance \cite{paetzold-specia-2017-lexical}, with the embedding of the target complex word as candidate substitutions. These models outperformed a previous supervised approach that relied on a pre-existing lexicon \cite{Biran2011}. \cite{Glavas2015} achieved an accuracy of 68.2 on the LS-2012 dataset, \cite{paetzold-specia-2017-lexical} produced an F1-score of 0.171 on the BenchLS dataset, whereas \cite{Biran2011} attained an accuracy of 3.4 and an F1-score of 0.136 on each study's dataset respectively. However, only the word embedding model provided by \cite{paetzold-specia-2017-lexical} surpassed the performance of other previous approaches to SG.

In 2017, word embedding models were state-of-the-art for SG. Word embedding models, such as Word2Vec \cite{Mikolov2013a}, were used alongside more traditional approaches, such as querying a lexicon, or generating candidate substitutions based on certain rules \cite{PaetzoldSpecia2017_surveyLS}. Word embedding models conducted SG by converting potential candidate substitutions into vectors, hence word embeddings, and then calculating which of these vectors had the highest cosine similarity, or lowest cosine distance, with the vector of the target complex word. These vectors were then converted back into their word forms and were considered the top-k candidate substitutions. 

% \footnote{EASIER: \url{ https://data.mendeley.com/datasets/ywhmbnzvmx/2}}

% Exploration of Spanish Word Embeddings for Lexical Simplification - Alarcón, et al., 2021
\paragraph{\bf{Word Embeddings + LLMs}} Post 2017, word embedding models continued to be implemented for SG. However, they were now combined with the word embeddings produced by LLMs or by a LLM's prediction scores. \citet{Alarcn2021ExplorationOS} experimented with various word embeddings models for generating Spanish candidate substitutions. They used word embeddings models, such as Word2Vec, Sense2Vec \cite{sense2vec}, and FastText \cite{fasttext}, along with the pre-trained LLM BERT, to generate these word embeddings. It was discovered that a more traditional approach that produced candidate substitutions by querying a pre-existing lexicon outperformed these word embedding models in terms of both potential and recall yet slightly under-performed these word embedding models in regards to precision. The traditional approach achieved a potential of 0.898, a recall of 0.597, and a precision of 0.043 on the EASIER dataset \cite{Alarcn2021LexicalSS}. The highest performing word embedding model (Sense2Vec), on the other hand, attained a potential, recall, and precision score of 0.506, 0.282, and 0.056, respectively. Surprisingly, this went against the assumption that word embedding models would have achieved a superior performance given their state-of-the-art reputation demonstrated by \citet{paetzold-specia-2017-lexical}. During error analysis, it was found that these word embeddings models often produced antonyms of the target complex word as potential candidate substitutions. This is due to how word embedding models calculate word similarity between vectors. 

% RCML as well
% CILS and 
% Arefyev (2020) - Always Keep your Target in Mind: Studying Semantics and Improving Performance of Neural Lexical Substitution 
% Zhou, et al., 2019 - BERT-based Lexical substitution

% on the TSAR-2022 dataset. \footnote{TSAR-2022: \url{https://taln.upf.edu/pages/tsar2022-st/}}

\citet{cils-tsar-2022-shared-task} used a word embedding model and a pre-trained LLM: XLNet \cite{yang_etal_2019}, to produce an embedding similarity score and a prediction score for SG. They followed a similar approach conducted by \citet{arefyev-etal-2020-always}. \citet{arefyev-etal-2020-always} utilized context2vec \cite{melamud-etal-2016-context2vec} and ELMo \cite{peters-etal-2018-deep} to encode the context of the target complex word to gain a probability distribution of each word belonging to that particular context. They then used this probability distribution to estimate the likelihood, or appropriateness, of a potential candidate substitution replacing the target complex word. This score was used alongside a LLM prediction score from either BERT, RoBERTa, or XLNet, to produce a final list of top-k candidate substitutions. Both \citet{cils-tsar-2022-shared-task} and \citet{arefyev-etal-2020-always} discovered that their combined approach of using a word embedding model alongside a pre-trained LLM prediction score failed to surpass the performance of using a single pre-trained LLM. For instance, \citet{cils-tsar-2022-shared-task} was outperformed by \citet{gmu-wlv-tsar-2022-shared-task} on the TSAR-2022 dataset.

\paragraph{\bf{Masked Language Modeling}} The introduction of pre-trained LLMs, also saw the arrival of Masked Language Modeling (MLM) for SG. \citet{przybyla-shardlow-2020-multi} used LLMs trained on a MLM objective for multi-word LS, whereas \citet{qiang2020BERTLS} were the first to use MLM for Spanish SG. MLM has subsequently become a popular approach to SG. 7 out of the 11 system reports submitted to TSAR-2022 \cite{tsar2022}, described their approach as consisting of a MLM objective.

 % the LS datasets LexMTurk\footnote{LexMTurk: \url{https://cs.pomona.edu/~dkauchak/simplification/}} \cite{Horn2014}, BenchLS\footnote{BenchLS: \url{https://zenodo.org/record/2552393\#.Y9Ku2nbMLcs}} \cite{Paetzold2016a}, and NNSeval\footnote{NNSeval: \url{https://zenodo.org/record/2552381\#.Y9Kuq3bMLcs}} 

Known as LSBert, the model introduced by \citet{qiang2020BERTLS}, used the pre-trained LLM BERT. Sentences were taken from the LS datasets LexMTurk\cite{Horn2014}, BenchLS \cite{Paetzold2016a}, and NNSeval \cite{Paetzold2016}. Two versions of each sentence were then concatenated, being separated by the [SEP] special token. They were then fed into the LLM. The first sentence was identical to that extracted from the datasets, whereas the second sentence had its complex word replaced with the [MASK] special token. The LLM then attempted to predict the word replaced by the [MASK] special token by taking into consideration its left and right context as well as the prior original sentence. In this way, LLMs provide candidate substitutions with the highest probability (highest prediction score) of fitting into the surrounding context and that are also similar to the target complex word in the original sentence. For the top-k=1 candidate substitution, LSBert achieved F1-scores for SG of 0.259, 0.272, and 0.218 on the three datasets LexMTurk \cite{Horn2014}, BenchLS \cite{Paetzold2016a}, and NNSeval \cite{Paetzold2016} respectively. These performances surpassed that of all prior approaches \cite{PaetzoldSpecia2017_surveyLS}. The previous highest F1-score was achieved by a word-embedding model \cite{paetzold-specia-2017-lexical}, which produced F1-scores of 0.195, 0.236, and 0.218 for each dataset, respectively.

% \paragraph{\bf{Monolingual MLM}}

Before the release of the TSAR-2022 shared-task \cite{tsar2022}, \citet{Ferres_Saggion2022} introduced a new dataset: ALEXSIS (TSAR-2022 ES), that would later make up (along with an additional English and Portuguese dataset) the TSAR-2022 dataset \cite{tsar2022}. Using their Spanish dataset, they experimented with a number of monolingual LLMs pre-trained on either Spanish data as well as several multilingual LLMs, such as mBERT and RoBERTa. \citet{Ferres_Saggion2022} adopted the MLM approach used by LSBert. They experimented with the Spanish LLMs: BETO \cite{CaneteCFP2020}, BERTIN \cite{BERTIN-GPT}, RoBERTa-base-BNE, and RoBERTA-large-BNE \cite{RoBERTa_es} for SG. They discovered that their largest pre-trained Spanish LLM: RoBERTA-large-BNE, achieved the greatest SG performance after having also removed candidate substitutions equal to the complex word, regardless of capitalization or accentuation and being less than 2 characters long.

%GMU-WLV - Monolingual MLM & PresiUniv
\citet{gmu-wlv-tsar-2022-shared-task} was inspired by the success of the monolingual LLMs shown by \citet{Ferres_Saggion2022}.  They likewise tested a range of LLMs for SG with a MLM objective, including multilingual LLLMs: mBERT, and XLM-R \cite{conneau-etal-2020-unsupervised}, and several monolingual LLMs, including Electra for English \cite{clarketal2020}, RoBERTA-large-BNE for Spanish, and BERTimbau \cite{souza2020bertimbau} for Portuguese. Their monolingual LLMs scored an acc@1 score of 0.517, 0.353, and 0.481 on the English, Spanish, and Portuguese TSAR-2022 datasets respectively. \citet{presiuniv-tsar-2022-shared-task} also experimented with similar monolingual LLMs for SG. They used BERT for English, BETO for Spanish, and BERTimbau for Portuguese. Interestingly, their models' performances were lower compared to that of \citet{gmu-wlv-tsar-2022-shared-task}, despite their Portuguese LS system consisting of the same language model. \citet{presiuniv-tsar-2022-shared-task} achieved acc@1 scores of 0.378, 0.250, and 0.3074 for English, Spanish, and Portuguese, respectively. This is likely due to the additional SS and SR steps implemented by \citet{presiuniv-tsar-2022-shared-task} and the lack thereof shown within the LS system provided by \citet{gmu-wlv-tsar-2022-shared-task} (Section \ref{SR}).

% 4). CENTAL - explored the use of masked language model for candidate generation with three strategies for context expansion:Copy, Query Expansion, and Paraphrase. The Copy strategy is a copy of the sentence itself (follows that of LSBert). The Query Expansion strategy extracts alternative words for the target word from FastText and then replaces the original sentence with each alternative word. The Paraphrase strategy (English only) extracts paraphrases from Pegasus (Zhanget al., 2020).

\citet{cental-tsar-2022-shared-task} also used a range of monolingual LLMs for SG. However, they used an ensemble of BERT-like models with three different masking strategies: 1). copy, 2). query expansion, and 3). paraphrase.  The copy strategy replicated that of LSBert \cite{qiang2020BERTLS}, whereby two sentences were inputted into a LLM concatenated with the [SEP] special token. The first sentence being an unaltered version of the original sentence, and the second sentence having its complex word masked. The query expansion strategy used FastText to generate five related words with the highest cosine similarity to the target complex word. For iteration 2a). of the query expansion strategy, the first sentence was the original unaltered sentence, the second sentence replaced the complex word with one of the suggested similar words produced by FastText, and sentence 3 was the masked sentence. Iteration 2b). of this strategy was the same as iteration 2a)., however, sentence 2 now consisted of all five suggested words. Lastly, the paraphrase strategy generated 10 new contexts for each complex word composed of paraphrases of the original sentence. These new contexts were limited to 512 tokens.  The ensembles used for these three masking strategies consisted of BERT and RoBERTa LLMs for English, several BETO LLMs for Spanish, and several BERTimbau LLMs for Portuguese. The paraphrase strategy showed the worst performance with a joint MAP/Potential@1 score of 0.217, whereas the query expansion strategy obtained a MAP/Potential@1 score of 0.528, 0.477, and 0.476 for English, Spanish, and Portuguese, respectively. This surpassed the performance of the paraphrase strategy and the original copy strategy used by LSBert, regardless of the LLMs used.

%%% GOT TO HERE &&&

% 2). UoM\&MMU - uses an approach that consists of three steps: 1) candidate generation based on different prompt templates (e.g., <easier, simple> <word, synonym> for <target_word>)

% For example, given the following template: ``A$<$Prompt1$><$Prompt2$>$for$<$target\_word$>$is'', two words were used for prompt 1). ``\textit{easier}'' and ``\textit{simple}'', and two other words were exchanged for prompt 2). ``\textit{word}'' and ``\textit{synonym}''. This resulted in four different input combinations:

\paragraph{\bf{Prompt Learning}} Prompt learning has also been used for SG and is currently state-of-the-art (Table \ref{deep_learning}). Prompt learning involves feeding into a LLM input that is presented in such a way as to provide a description of the task as well as to return a desired output. PromptLS is an example of prompt learning applied to SG. Created by \citet{uom-mmu-tsar-2022-shared-task}, PromptLS consisted of a variety of pre-trained LLMs fine-tuned on several LS datasets. These fined-tuned LLMs were then presented with four combinations of prompts: a). ``a \underline{\textit{easier}} \underline{\textit{word}} for bombardment is'', b). ``a \underline{\textit{simple}} \underline{\textit{word}} for bombardment is'', c). ``a \underline{\textit{easier}} \underline{\textit{synonym}} for bombardment is'', and lastly, d). ``a \underline{\textit{simple}} \underline{\textit{synonym}} for bombardment is''. These prompt combinations were supplied to a RoBERTa LLM on all of the English data extracted from the LexMTurk \cite{Horn2014}, BenchLS \cite{Paetzold2016a}, NNSeval \cite{Paetzold2016}, and CERF-LS \cite{uchida2018} LS datasets. They were also translated and fed into BERTIN fine-tuned on the Spanish data obtained from EASIER, along with BR-BERTo fine-tuned on all of the Portuguese data taken from SIMPLEX-PB \cite{Hartmann2020}. \citet{uom-mmu-tsar-2022-shared-task} also used these prompts on a zero-shot condition. It was discovered that the fine-tuned LLMs outperformed the zero-shot models on all conditions by an average increase in performance between 0.3 to 0.4 across all metrics: acc@1, acc@3, MAP@3, and Precision@3. The prompt combinations that produced the best candidate  substitutions were ``easier word'' for English, ``palabra simple''  and ``palabra fácil'' for Spanish, and ``palavra simples''  and ``sinônimo simples'' for Portuguese.

% 1). UniHD - GPT-3.5 
Prompt learning has likewise been applied to causal language models for SG, such as GPT-3. \citet{unihd-tsar-2022-shared-task} experimented with a variety of different prompts, which they fed into a GPT-3. These prompts were of four types: 1). zero-shot with context, 2). single-shot with context, two-shot with context, 3). zero-shot without context, and 4). single-shot without context. The size of each shot: \textit{n}, refers to how many times a prompt is inputted into GPT-3. For instance, those shots with context would input a given sentence and then ask the question, ``Given the above context, list ten alternative words for $<$complex word$>$ that are easier to understand.'', \textit{n} number of times. Those without context, however, would input \textit{n} times the following:``Give me ten simplified synonyms for the following word: $<$complex word$>$''. \citet{unihd-tsar-2022-shared-task} also combined all types of prompts in an ensemble, generating candidate substitutions from each prompt type and then deciding upon final candidate substations through plurality voting and additional SS and SR steps (Section \ref{SR}). Their ensemble approach outperformed all other prompt types and SG models submitted to TSAR-2022 \cite{tsar2022} (Table \ref{deep_learning}).

% \subsection{Substitute Selection} \label{SS}

\subsection{Substitute Selection and Ranking} \label{SR}

%  Mentioned in prior survey - Paetzold and Specia, 2017 - Lexical Simplification with Neural Ranking
% Start with intro explain where Paetzold and Specia left off for SS and SR. Refer to the above study.
% Also mentionn how SS and SR and normally conducted simultaneously and that selection techniques, when on their own, have not much changed since 2017.

Traditional approaches to SS are still implemented post SG. Methods such as POS-tag and antonym filtering, semantic or sentence thresholds have been used to remove inappropriate candidate substitutions after having been generating from the above deep learning approaches \cite{tsar2022}. Nevertheless, the majority of modern deep learning approaches have minimal SS, with SS often being simultaneously conducted during SG or SR. For instance, the metric used to generate the top-k candidate substitutions, by it either similarity between word embeddings, or a pre-train LLM's prediction score, tends not to suggest candidate substitutions that are deemed as being inappropriate by other SS methods. Likewise, SR techniques that rank candidate substitutions in order of their appropriateness will in turn move inappropriate simplifications further down the list of top-k candidate substitutions to the point that they are no longer considered.

%CILS - 1) the score from the candidate generation; 2) sentence similarity score (cosine similarity between the source and target sentence); 3) gloss sentence similarity score (the cosine similarity between the target word and the candidate);
% \paragraph{\bf{Semantic Similarity}} 

 %using WordNet \cite{miller1995wordnet}
% Song, et al., 2020 - Personalized Lexical Simplifcation
\paragraph{\bf{Word Embeddings}} Word embedding models continued to be used for SS without LLMs, regardless of the arrival of pre-trained LLMs, such as BERT. For instance, \citet{song_et_al_2020} created a unique LS system that filtered candidate substitutions by applying a semantic similarity threshold, matching only those  candidate substitutions with the same POS tag as the target complex word, calculating contextual relevance, being a measure of how reasonable and fluent a sentence is after the complex word had been replaced, and by using cosine similarity between word embeddings to rank candidate substitutions. They generated word embeddings by Word2Vec and evaluated their model's performance on the LS-2007 dataset \cite{mccarthy-navigli-2007-semeval}. It was found that the use of Word2Vec improved their model's performance having achieved an acc@1 of 0.269. Their second highest performing model, without the use of Word2Vec embeddings, produced an acc@1 of 0.218.

\paragraph{\bf{Neural Regression}}
% Maddela and Xu, 2018 - WCL - A Word-Complexity Lexicon and A Neural Readability Ranking Model for Lexical Simplification. Evaluated on the LS-2012 dataset.

% The magnitude of the returned value represents the degree of difference between the two candidate substitutions \cite{north-etal-2022-evaluation}

% Word Complexity Lexicon\footnote{WCL: \url{https://github.com/mounicam/lexical_simplification}}

\citet{maddela2018word} created the neural readability ranker (NNR) for SR. Consisting of a feature extraction, a Gaussian-based feature vectorization layer, and a task specific output node, NNR is a deep learning algorithm capable of ranking candidate substitutions based on their perceived complexity. It performances regression, whereby having been trained on the Word Complexity Lexicon (WCL), as well as several features and character n-grams converted into Gaussian vectors, it is able to provide a value between 0 and 1 corresponding to the complexity of any given word. It achieves this by conducting pairwise aggregation. For each pair of potential candidate substitutions, the model predicts a value that defines which candidate substitution is more or less complex than the other. A return positive value indicates that the first candidate substitution is more complex than the second, whereas a negative value dictates that the second candidate substitution is more complex than the first. This is applied to all combinations of candidate substitutions given a complex word. Each candidate substitution is then ranked in accordance to its comparative complexity with all other potential candidate substitutions. \citet{maddela2018word} applied their NNR model to the LS-2012 dataset and outperformed prior word embedding techniques for SR. They achieved an Prec@1 of 0.673, whereas the previous state-of-the-art model provided by \citet{paetzold-specia-2017-lexical} achieved an Prec@1 of 0.656.

% . The second and third ranking metrics made use of  WordNet \cite{miller1995wordnet} definitions of the target 

\paragraph{\bf{Word Embeddings + LLMs}} One of the most common approaches to SS and SR involves the use of word embeddings and LLMs. \citet{cils-tsar-2022-shared-task} filtered and ranked top-k=20 candidate substitutions based on the same combined score that they used for SG.  It consisted of their MLM model's prediction score of the generated candidate together with the inner product of the target word's embedding and the embedding of the potential candidate substitution. These top-k=20 candidate substitutions were then subject to one of three additional ranking metrics. The first ranking metric (CILex\_1) ranked candidate substitutions on their cosine similarity between the original sentence and a copy of the original sentence with the candidate substitution in place of its complex word. The second and third ranking metrics made use of dictionary definitions of the target complex word and its candidate substitutions. They calculated the cosine similarity between each embedding of each definition and the embedding of the sentence of the target complex word. Those with the highest cosine similarities between a). the definition of the target complex word and the definition of the candidate substitution (CILex\_2), or b). the definition of the target complex word and the word embedding of the original sentence with the candidate substitution in place of its complex word (CILex\_3), were used to determine the rank of each candidate substitution. They discovered that all three metrics produced similar performances on the TSAR-2022 dataset with CILex 1, 2, and 3 achieving acc@1 scores of 0.375, 0.380, and 0.386, respectively. 

% \begin{equation}
% S_{XLNet} = \alpha P(w|c) + \beta P(w|x)
% \end{equation}\label{ranking_CILS2}

% \cite{} (MANTIS) used RoBERTa \cite{} to generate candidate substitutions for English. They increased the number of candidate substitutions generated by their model to 30, rather than the 10 outputted by LSBert. They hoped that after having implemented additional steps in SS and SR (Section \ref{SR}), this increase in the number of candidates generated would also increase the probability of capturing at least one of the gold label simplifications within the TSAR-2022 dataset. Their system came 2nd on the English track at TSAR-2022 \cite{} having achieved an acc@1 of 0.8096.  

%MANTIS - semantic similarity (cosine similarity between the FastText vector of the target word and the candidate). Counter to using word embeddings.

% The model was trained on the Multi-Genre Natural Language Inference corpus \cite{williams-etal-2018-broad} 

\citet{mantis-tsar-2022-shared-task} used a set of features taken from LSBert combined with what they referred to as an equivalence score. Equivalence score was created to gauge semantic similarity between candidate substitution and complex word to an extent that was more expressive than the cosine similarity between word embeddings. To obtain this equivalence score, they used a pre-trained RoBERTa LLM trained for natural language inference (NLI) which predicts the likelihood of one sentence entailing another. The model was trained on a multi-genre corpus with a MLM objective. The product of the returned likelihood of the original sentence with the candidate substitution preceding the original sentence and vice-versa equated to the equivalence score. Since \citet{mantis-tsar-2022-shared-task} used the same method of SG as LSBert, having only changed their LLM to RoBERTa, they concluded that their system's superior performance was a consequence of its unique SR. They achieved an acc@1 of 0.659, whereas LSBert attained an acc@1 of 0.598 on the English TSAR-2022 dataset \cite{tsar2022}.

\citet{rcml-tsar-2022-shared-task}  ranked candidate substitutions on three metrics: a). grammaticality, b). meaning preservation, and c). simplicity. Grammaticality was calculated by firstly determining whether the candidate substitution had the same POS tag in terms of person, number, mood, tense, and so forth. Those that matched on all POS-tag categories were assigned the value of 1 or 0 if at least one category did not match. Preservation was determined by using BERTScore to generate cosine similarities between the embeddings of the original sentence and the embeddings of the original sentence, having replaced the target complex word with the candidate substitution. Lastly, preservation was obtained by using a CEFR vocabulary classifier trained on data from the English Vocabulary Profile (EVP).  The data used to train the CEFR classifier was first masked and fed into a pre-trained LLM: BERT. The outputted encodings were then used to train an SVM model resulting in their CEFR classifier. Their model failed to surpass the baseline LSBert models at TSAR-2022 in terms of acc@1, having achieved a score of 0.544. 

% However, upon further examination they discovered 15 instances within the TSAR-2022 dataset whose gold top-k=1 candidate substitution was seen as being more complex that the top-k=1 candidate substitution supplied by their model. 

% GMU-WLV also
% Both GMU-WLv and UoM\&MMU both used the LM prediction score for ranking. - Combine these in one paragraph!
\paragraph{\bf{MLM Prediction Scores}} LS systems have also relied entirely on MLM prediction scores for SS and SR. \citet{gmu-wlv-tsar-2022-shared-task} and \citet{uom-mmu-tsar-2022-shared-task} adopt this approach. They have no additional SR steps and rank their candidate substitutions per their generated MLM prediction scores. They do, however, apply some basic filtering with both studies removing duplicates as well as candidate substitutions equal to the complex word. Surprisingly, minimal SR has been shown to surpass other more technical approaches (Table \ref{deep_learning}). \citet{gmu-wlv-tsar-2022-shared-task} has achieved state-of-the-art performance on the TSAR-2022 Portuguese dataset, whereas \citet{uom-mmu-tsar-2022-shared-task} has consistently produced high performances across the English and Spanish TSAR-2022 datasets. Only GPT-3 based-models have surpassed these performances \cite{unihd-tsar-2022-shared-task} (Table \ref{deep_learning}).

\section{Resources}\label{datasets} 

%  !). Put datasets in table with citations. Datasets then Benchmark Competitions

% 2). In the preceding sections (4-7), discuss the best performing systems, only those which have deep learning models; also, preferably after 2017. 

% ===============================================
%==================== FORMAT ====================
% ===============================================

% 1. English
    % - English datasets post 2017 that have yet to be described.
    
% 2. Other Language
    % - Spanish
    % - Portuguese
    % - French
    % - Japanese
    % - Chinese

% Table \ref{datasets_table} (Appendix).shows datasets (including CWI datasets) that can be used for LS. It summarizes the sub-tasks that they can be used for, as well as their language(s), size, domain, and annotators.

Post 2017 LS datasets have been created for either all sub-tasks within the LS pipeline or for a specific purpose (Appendix, Table \ref{datasets_table}). Recent international competitions (shared-tasks) have also provided their own LS datasets (*). LS resources are available for multiple languages, predominately English (EN), Spanish (ES), Portuguese (PT), French (FR), Japanese (JP), and Chinese (ZH). 

% Post 2017 datasets have been grouped per their respective language and described in the following sections. 

\subsection{English}\label{en_datasets}

\paragraph{\bf{Personalized-LS}} \citet{lee-yeung-2018-personalizing} constructed a dataset of 12,000 English words for personalized LS. These words were ranked on a five-point Likert scale. 15 native Japanese speakers were tasked with rating the complexity of each word. These complexity rating were then applied to BenchLS, in turn personalizing the dataset for Japanese speakers.

% The dataset can be used for SG, SS and SR since it provides continuous complexity ratings.

\paragraph{\bf{WCL}} \citet{maddela2018word} introduced the Word Complexity Lexicon (WCL). The WCL is a dataset made up of 15,000 English words annotated with complexity ratings. Annotators were 11 non-native English speakers using a six-point Likert scale.

% The dataset can be used for SS and SR.

% \footnote{CompLex: \url{https://github.com/MMU-TDMLab/CompLex}}
 % Citation Network Dataset\footnote{SimpleText-2021: \url{https://www.aminer.org/citation}}
 
\paragraph{\bf{LCP-2021*}}
The dataset provided at the LCP-2021 shared-task (CompLex) \cite{shardlow-etal-2020-complex}, was developed using crowd sourcing. 10,800 complex words in context were selected from three corpora covering the Bible, biomedical articles, and European Parliamentary proceedings. Their lexical complexities were annotated using a 5-point Likert scale. 

% This is another useful dataset for SS and SR.

\paragraph{\bf{SimpleText-2021*}} The SimpleText-2021 shared-task \cite{SimpleText_2021} introduced three pilot tasks: 1). to select passages to be simplified, 2). to identify complex concepts within these passages, and 3). to simplify these complex concepts to generate an easier to understand passage. They provided their participants with two sources of data, these being the  Citation Network Dataset, DBLP+Citation, ACM Citation network, together with titles extracted from The Guardian newspaper with manually annotated keywords. 

% Portuguese \cite{north2022alexsis}

\paragraph{\bf{TSAR-2022*}}
TSAR-2022 \cite{tsar2022} supplied datasets in English, Spanish, and Portuguese. These datasets contained target words in contexts taken from journalistic texts and Wikipedia articles, along with 10 candidate substitutions (approx. 20 in raw data) provided by crowd-sourced annotators located in the UK, Spain, and Brazil. The candidate substitutions were ranked per their suggestion frequency. The English, Spanish, and Portuguese datasets contained 386, 381, and 386 instances, respectively.

\subsection{Datasets in Other Languages}\label{other_datasets}

% \subsection{Spanish}\label{es_datasets}

 \paragraph{\bf{Spanish}} The ALexS-2020 shared-task \cite{Zambrano2020OverviewOA} included a Spanish dataset consisting of 723 complex words from recorded transcripts. \citet{merejildo2021} provided the Spanish CWI corpus (ES-CWI). A group of 40 native-speaking Spanish annotators identified complex words within 3,887 academic texts. The EASIER corpus \cite{Alarcn2021LexicalSS} contains 5,310 Spanish complex words in contexts taken from newspapers with 7,892 candidate substitutions. A small version of the corpus is also provided with 500 instances (EASIER-500). 
 
 % Both the ALexS-2020 and ES-CWI datasets can be used for SG. The EASIER corpus can be used for SG and SS, but lacks the necessary ranking of candidate substitutions for SR.

 % EASIER dataset and EASIER-500 dataset

% \subsection{Portuguese}\label{pt_datasets}

 % newest version of the SIMPLEX-PB dataset \cite{hartmann-etal-2020-simplex_a}.

\paragraph{Portuguese} The PorSimples dataset \cite{Aluisio2010} consists of extracts taken from Brazilian newspapers. The dataset is divided into nine sub-corpora separated by degree of simplification and source text. The PorSimplesSent dataset \cite{leal-etal-2018-nontrivial} was adapted from the previous PorSimples dataset. It contains strong and natural simplifications of PorSimples's original sentences. SIMPLEX-PB \cite{Hartmann2020} provides a selection of features for each of its candidate substitutions. 

% The PorSimples and PorSimplesSent datasets can be used for SG and SS, whereas SIMPLEX-PB can also be used for SR.

%ALECTOR\cite{gala-etal-2020-alector}

\paragraph{\bf{French}} ReSyf contains French synonyms that have been ranked in regards to their reading difficulty using a SVM \cite{billami-etal-2018-resyf}. It consists of 57,589 instances with a total of 148,648 candidate substitutions. FrenchLys is a LS tool designed by \citet{rolin-etal-2021-frenlys}. It provides its own dataset that contains sentences sampled from a French TS dataset: ALECTOR, and french schoolbooks. Substitute candidates were provided by 20 French speaking annotators. 

% Both datasets are excellent resources for French SG, SS, and SR.

% \subsection{Japanese}\label{jp_datasets}

%from the Balanced Corpus of Contemporary Written Japanese \cite{maekawa-etal-2010-design}

\paragraph{\bf{Japanese}} The Japanese Lexical Substitution (JLS) dataset \cite{kajiwara-yamamoto-2015-evaluation} contains 243 target words, each with 10 contexts (2,430 instances in total).  Crowd-sourced annotators provided and ranked candidate substitutions. The JLS Balanced Dataset \cite{kodaira-etal-2016-controlled} expanded the previous JLS dataset to make it more representative of different genres and contains 2,010 generalized instances. \citet{nishihara-kajiwara-2020-word} created a new dataset (JWCL \& JSSL) that increased the Japanese Education Vocabulary List (JEV). It houses 18,000 Japanese words divided into three levels of difficulty: easy, medium, or difficult.  

% The JLS datasets can be used for all sub-tasks within the LS pipeline. However, the dataset provided by  \citet{nishihara-kajiwara-2020-word} is more suited for SG.

% \footnote{HanLS: \url{https://github.com/luxinyu1/Chinese-LS}}

\paragraph{\bf{Chinese}} Personalized-ZH \cite{Lee_Yeung2018} consists of 600 Chinese words. Each word's complexity was ranked by eight learners of Chinese on a 5-point lickert-scale. HanLS was constructed by \citet{HanLS}. It contains 534 Chinese complex words. 5 native-speaking annotators gave and ranked candidate substitutions. Each complex word has on average 8 candidate substitutions.

\section{Discussion and Conclusion}\label{conclusion}

Since the 2017 survey on LS \cite{PaetzoldSpecia2017_surveyLS}, deep learning approaches have provided new headway within the field. MLM is now the go to method for SG, with the majority of recent LS studies having employed a MLM objective. The casual language model: GPT-3, surpasses the performance of all other approaches when subjected to prompt learning, especially when an ensemble of prompts are taken into consideration (Table \ref{deep_learning}). The prediction scores of MLM or casual language modeling have replaced various SS and SR techniques. LS systems that employ minimal SS and no SR apart from ranking their LLM's prediction scores, have outperformed more technical, feature-oriented, and unsupervised ranking methods (Table \ref{deep_learning}). However, an exception is made with regards to equivalence score \cite{mantis-tsar-2022-shared-task}, which has been shown to be effective at SR. 

Future LS systems will make use of new advances in deep learning. We believe  prompt learning and models, such as GPT-3, will become increasingly popular, given their state-of-the-art performance at SG. Using an ensemble of various prompts for SS and SR may advance LS performance. In addition, the creation of new metrics similar to equivalence score will likewise be beneficial.

% \subsection{State-of-the-Art}

\subsection{Open Challenges in LS}\label{challenges}

LS has a number of open research areas that are either unaddressed, or the current body of work is inconclusive. In this brief section, we conclude this survey by outlining a few key areas for future development of LS research.

\paragraph{Evaluation:} The metrics we use to evaluate LS are not perfect (Section \ref{eval}). Automated metrics that condense a wide problem into a single numerical score can harm outcomes with human participants. Development of more faithful resources, as well as direct evaluation with intended user groups of simplification systems is a fruitful avenue for future work. This can be done by taking into consideration variation in data annotation instead of labels produced by aggregating unique annotations as in most datasets currently available.

\paragraph{Explainability:} Lexical simplifications are inherently more explainable than sentence simplification as the operations are directly applied at the lexeme level. However, the decision process on whether to simplify and which word to choose is increasingly hidden behind the black-box of a model. Work to explain and interpret these decisions will allow researchers to better understand the opportunities and threats of applying  modern NLP techniques to LS research.

\paragraph{Personalization:} One model does not fit all. The simplification needs of a language learner compared to a stroke victim, compared to a child are each very different. Modeling these needs and using them to personalize LS systems will allow for personalized simplification output more adequate the needs of particular user groups.

\paragraph{Perspectivism:} Even within a population of common characteristics, each individual will bring a unique perspective on what and how to simplify. Systems which can alter their outputs to each user's needs will provide adaptive simplifications that go beyond our current technology. This will, in turn, improve the evaluation of LS models as previously discussed in this section.

\paragraph{Integration:} LS is only one part of the wider simplification puzzle. Integrating LS systems with explanation generation, redundancy removal, and sentence splitting will further accelerate the adoption of automated simplification practices beyond the halls of research allowing such technology to reach a wider audience.

\bibliography{custom, CWI}
\bibliographystyle{acl_natbib}

\appendix

\begin{landscape}

\section{Appendix}
\label{sec:appendix}

% ----/ English CWI/LCP Datasets Table /---- 
\begin{table}[!ht]
\centering
\scalebox{1}{\begin{tabular}{cp{3cm}cccccp{4cm}c}
\hline
  & \bf Dataset & \bf LS Pipeline & \bf Languages & \bf \# CWs & \bf Avg. \# Subs & \bf Domain & \bf Annotators  & \bf Paper  \\ \hline

\multirow{10}{*}{\rotatebox[origin=c]{90}{Pre-2017}} & \bf   LS--2007* & SG, SS & EN & 201 & 1 & Mix & 5 UK-based. & \cite{mccarthy-navigli-2007-semeval} \\

& \bf   PorSimples & SG, SS & PT & 3066 & 1 & News & 1 Linguist. & \cite{Aluisio2010}\\

 & \bf   LS--2012* & SG, SS, SR & EN & 201 & 5 & Mix & L1 English Speakers. & \cite{specia2012} \\

& \bf   CW Corpus & SS & EN & 731 & 0 & Wikipedia & Wikipedia Edits.& \cite{CWcorpus}  \\ 

& \bf   LexMTurk & SG, SS, SR & EN & 500 & 50 &  Wikipedia & 50 US-based. & \cite{Horn2014}\\

& \bf   JLS & SG, SS, SR & JP & 243 & 5 & Mix & 5 L1 JP Speakers. & \cite{kajiwara-yamamoto-2015-evaluation}   \\

& \bf   JLS Balanced & SG, SS, SR & JP & 2,010 & 5 & Mix & L1 JP Speakers & \cite{kodaira-etal-2016-controlled}  \\

& \bf   CWI--2016* & SS & EN & 90,458 & 0 & News & 400 L2 EN Speakers. & \cite{paetzold-specia:2016:SemEval1}  \\ 

& \bf   BenchLS & SG, SS, SR & EN & 929 & 7 & Mix & US-Based. & \cite{Paetzold2016a} \\

& \bf   NNSeval & SG, SS, SR & EN & 239 & 7 & Mix & 400 L2 EN Speakers. & \cite{Paetzold2016} \\

\hline

\multirow{17}{*}{\rotatebox[origin=c]{90}{Post-2017}}  & \bf   CERF-LS & SG, SS, SR & EN & 406 & 12 & Academic & 1 L1 EN Speaker. &  \cite{uchida2018}\\

& \bf   Personalized-ZH & SG, SS, SR & ZH & 600 & 7 & Mix & 8 L1 ZH Speakers & \cite{Lee_Yeung2018} \\

& \bf   WCL & SS, SR & EN & 15,000 & 0 & Mix & 11 L2 EN Speakers.  & \cite{maddela2018word} \\ 

& \bf   ReSyf & SG, SS & FR & 57,589 & 3 & Mix & L1 FR Speakers. & \cite{billami-etal-2018-resyf} \\

& \bf   Personalized-LS & SG, SS, SR & EN & 929 & 7 & Mix & 15 L2 EN Speakers. & \cite{lee-yeung-2018-personalizing} \\

& \bf   CWI--2018* &  SS, SR & EN, FR, GR, ES & 62,550 & 0 & News & L1\&L2 EN Speakers. & \cite{yiman-EtAl:2018:BEA} \\

& \bf   PorSimplesSent & SG, SS & PT & 6109 & 1 & News & 3 Linguists. & \cite{leal-etal-2018-nontrivial} \\

& \bf  LCP-2021* & SS, SR & EN &  10,800 & 0 & Mix & 7 US/UK/AUS-based. & \cite{shardlow-etal-2020-complex} \\

& \bf   SIMPLEX-PB & SG, SS, SR &  PT & 730 & 5 & Academic & pt-BR Speakers. & \cite{Hartmann2020} \\

& \bf   JWCL-JSSL & SG & JP & 18,000 & 0 & Mix & 5 L1 JP Speakers. & \cite{nishihara-kajiwara-2020-word} \\

& \bf   ALexS-2020* & SG & ES & 723 & 0 & Academic & 430 ES Speakers. & \cite{Zambrano2020OverviewOA} \\

& \bf   SimpleText-2021* & SG, SS, SR & EN & 1000 & 10 & Academic & Participating Teams. & \cite{SimpleText_2021} \\

& \bf   ES-CWI & SG  & ES & 3,887 & 0 & Academic & 40 L1 ES speakers. & \cite{merejildo2021} \\

& \bf   EASIER & SG, SS & ES & 5,310 & 3 & News & L1 ES speakers. & \cite{Alarcn2021LexicalSS} \\

& \bf   FrenLys & SG, SS, SR & FR & 57,589 & 3 & Mix & 20 L1 FR Speakers. & \cite{rolin-etal-2021-frenlys} \\

& \bf   HanLS & SG, SS, SR & ZH & 534 & 8 & Mix & 5 L1 ZH Speakers. & \cite{HanLS} \\

& \bf   TSAR-2022* & SG, SS, SR &  EN, ES, PT & 1153 & 20 & News & 21 UK/ES/BR-based. & \cite{tsar2022} \\

\hline

\end{tabular}}
\caption{Datasets that can be used for LS arranged in chronological order. Marked datasets (*) were used in benchmark competitions. L1 and L2 refers to first and second language speakers.}\label{datasets_table}
\end{table}

\end{landscape}

\end{document}